\documentclass[conference]{IEEEtran}
\IEEEoverridecommandlockouts
\usepackage{amsmath,amssymb,amsfonts}
\usepackage{graphicx}
\usepackage{textcomp}
\usepackage{booktabs}
\usepackage{makecell}
\usepackage{multirow}
\usepackage{url}
\usepackage{cite}
\usepackage{xcolor}
\usepackage{CJKutf8}
\usepackage[CJKbookmarks=true]{hyperref}

\renewcommand{\textbf}[1]{{\fontfamily{ptm}\fontseries{b}\selectfont #1}}
\renewcommand{\textit}[1]{{\fontfamily{ptm}\fontshape{it}\selectfont #1}}

\def\BibTeX{{\rm B\kern-.05em{\sc i\kern-.025em b}\kern-.08em
    T\kern-.1667em\lower.7ex\hbox{E}\kern-.125emX}}

\begin{document}

\title{FormalASR: End-to-End Spoken Chinese to Formal Text}

% \thanks{$^\ast$Corresponding author.}
\author{
  \IEEEauthorblockN{Wanyi Ning$^{1,2}$,
  Yinshang Guo$^3$,
  Haitao Qian$^1$,
  Jiyuan Cheng$^1$,
  Wei Zhou$^1$,
  Weiyuan Feng$^1$,
  Yufei Zhang$^1$}
  \IEEEauthorblockA{\textit{$^1$Yijiahe AI, Nanjing, China \quad
  $^2$Tianjin University, Tianjin, China \quad
  $^3$Nanjing University, Nanjing, China} }
  % ningwanyi@126.com, guoyinshang@gmail.com, 
  % \{chris, chengjiyuan, zhouwei, fengweiyuan, zhangyufei\}@yijiahe.com}
}

\maketitle

\begin{abstract}
Automatic speech recognition (ASR) systems are typically optimized for
verbatim transcription, which preserves disfluencies, filler words, and
informal spoken structures that are often unsuitable for downstream
writing-oriented applications.
A common workaround is a two-stage ASR+LLM pipeline for post-editing,
but this design increases latency and memory cost and is difficult to
deploy on-device.
We present FormalASR, two compact end-to-end models with 0.6B and
1.7B parameters that directly transcribe spoken Chinese into formal written text.
To enable this setting, we build WenetSpeech-Formal and
Speechio-Formal, two large-scale spoken-to-formal datasets
constructed by LLM-based rewriting and quality filtering.
We then fine-tune Qwen3-ASR at both model scales with
supervised fine-tuning.
Experiments on WenetSpeech-Formal and Speechio-Formal show that FormalASR achieves up
to 37.4\% relative CER reduction over verbatim baselines, while also
improving ROUGE-L and BERTScore.
FormalASR requires no post-processing LLM at deployment time, providing a lightweight, on-device solution for spoken-to-formal transcription.
\end{abstract}


% -------------------------------------------------------
\section{Introduction}
\label{sec:intro}
Automatic speech recognition (ASR) has become a foundational component
of modern human-computer interaction, powering applications ranging from
voice assistants and meeting transcription to real-time captioning and
document dictation\cite{malik2021automatic,sudo2024contextualized,hinton2012deep,graves2006ctc,chan2016listen}.
State-of-the-art systems such as Whisper~\cite{radford2023whisper},
Qwen3-ASR~\cite{qwen3asr2025}, and SenseVoice~\cite{sensevoice2024}
have achieved remarkable accuracy on standard benchmarks, yet they
share a fundamental design assumption: the output should faithfully
reproduce the spoken surface form.
This verbatim transcription paradigm faithfully captures what was said, but the output retains many characteristics of spontaneous speech, including filler words, false starts~\cite{zayats2016disfluency,jamshid2020improving}, repetitions, and disfluent structures~\cite{shriberg1994preliminaries,wang2020spoken}, which can reduce readability in formal written contexts.
In practice, however, many downstream applications require formal, well-structured text rather than verbatim transcripts that preserve the disfluencies of spontaneous speech.

\begin{figure}[t]
  \centering
  \includegraphics[width=\linewidth]{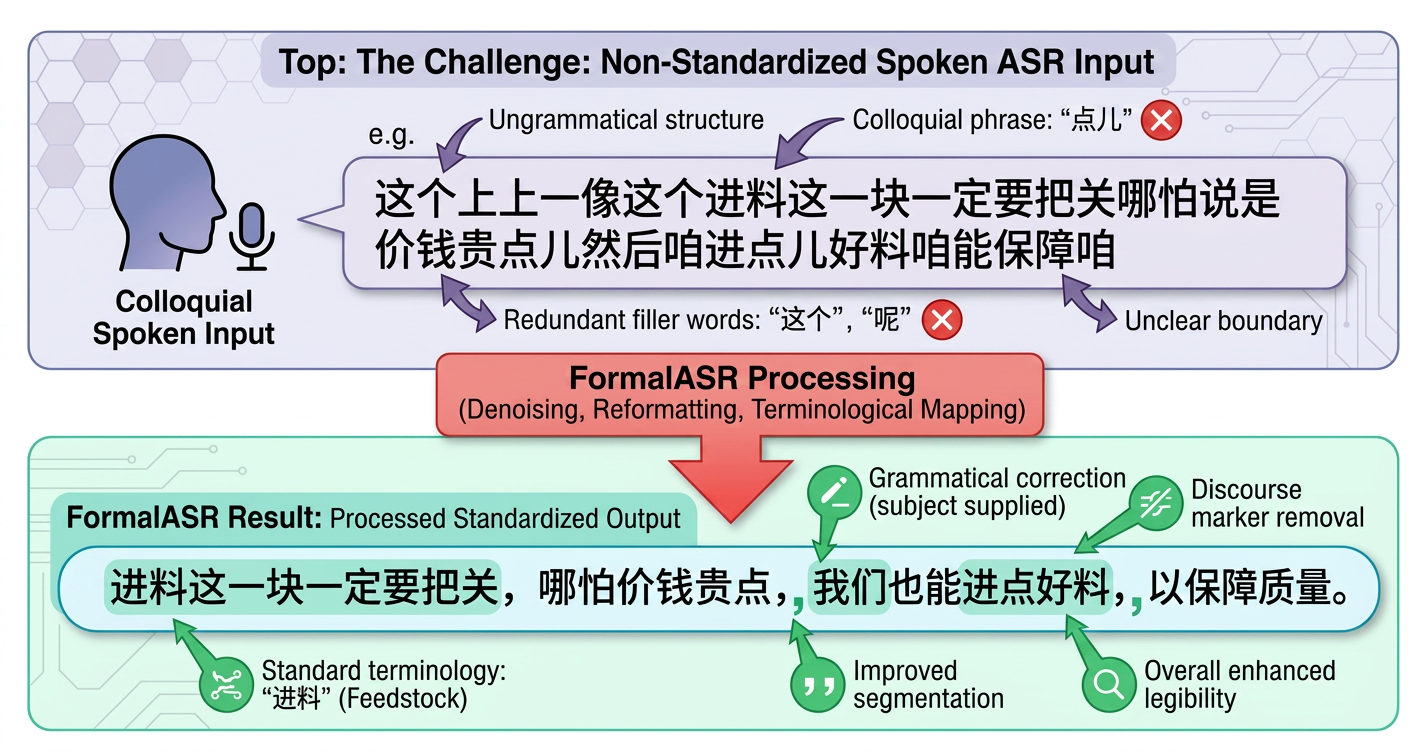}
  \caption{An example of spoken-to-formal conversion.
           The verbatim ASR transcript preserves disfluencies and
           informal spoken patterns, while FormalASR directly produces
           a clean, formal written sentence from the same audio input.}
  \label{fig:motivation}
\end{figure}

A common remedy is a two-stage pipeline: first transcribe verbatim
with an ASR model, then apply a separate large language model (LLM) to
rewrite the transcript into formal style~\cite{chen2023hyporadise,xu2025large,Sachdev2025evolutionary,yang2023generative}.
While effective in server-side settings, this approach increases
memory footprint and inference latency and allows errors to propagate
across stages. Large multimodal models such as GPT-4o-audio-preview
and Qwen3-Omni \cite{openai2024gpt4o,xu2025qwen3omni} can produce
formal-style transcriptions in a single pass, but depend on cloud
inference, incurring usage costs and raising privacy concerns that
limit deployment in on-device and resource-constrained settings.
As illustrated in Figure~\ref{fig:motivation}, the gap between
verbatim ASR output and the expected formal written form can be
substantial even for a single utterance. While prior work has studied spoken-to-written conversion at the text level, existing solutions typically operate on text transcripts produced by ASR systems, and no existing approach embeds this capability directly into an acoustic model~\cite{guo2023cs2w,wang2020spoken,sproat2001normalization,sproat2017rnn,choi2024spoken,liu2025recording}.

In this work, we propose FormalASR, an end-to-end approach
that directly maps spoken Chinese audio to formal written text using a
single compact model with no auxiliary LLM required at inference time.
The key insight is that an audio-language model can be taught to
perform acoustic recognition and linguistic formalization
simultaneously, provided it is trained on appropriate
spoken-to-formal supervision.
To supply this supervision at scale, we construct and open-source
WenetSpeech-Formal\footnote{\url{https://huggingface.co/datasets/TaurenMountain/WenetSpeech-Formal}}
and Speechio-Formal\footnote{\url{https://huggingface.co/datasets/TaurenMountain/Speechio-Formal}},
two large-scale Chinese spoken-to-formal ASR datasets derived from
WenetSpeech~\cite{zhang2022wenetspeech} and
Speechio~\cite{speechio2021}, built by rewriting verbatim
transcriptions with DeepSeek-V3.2~\cite{deepseek2024v3} and applying quality filtering.
We then fine-tune and open-source FormalASR\footnote{0.6B: \url{https://huggingface.co/TaurenMountain/FormalASR-0.6B};
1.7B: \url{https://huggingface.co/TaurenMountain/FormalASR-1.7B}},
compact audio--language models with two scales, 0.6B and 1.7B parameters,
using supervised fine-tuning (SFT) on the trainset of WenetSpeech-Formal.
Extensive experiments show that FormalASR achieves up to 37.4\% relative CER reduction with consistent ROUGE-L and BERTScore improvements over verbatim baselines on both
in-domain and cross-domain Chinese benchmarks, while requiring no
auxiliary model at inference time and remaining suitable for on-device deployment.

To the best of our knowledge, FormalASR is the first work that end-to-end fine-tunes a compact audio-language model to transcribe spoken Chinese into formal written Chinese.
Our results reveal that modern ASR models already possess the
latent capacity for linguistic formalization—they simply need
appropriate supervision to activate it, without any increase in
model size or inference-time complexity.

\begin{table}[t]
\centering
\caption{Representative spoken-to-formal conversion examples.}
\label{tab:data_examples}
\begin{CJK*}{UTF8}{gbsn}
\begin{tabular}{p{0.45\linewidth}p{0.45\linewidth}}
\toprule
\textbf{Verbatim (original\_text)} & \textbf{Formal (target\_text)} \\
\midrule
把这个呃增加的这个利润 & 把这个增加的利润。 \\
\midrule
对全美国全球影响影响不大 & 对美国全球影响不大。 \\
\midrule
但是我想这里这当中就是如果如果一定要那个挑一点儿什么的话 & 但是，如果一定要从中挑出一点什么的话。 \\
\bottomrule
\end{tabular}
\end{CJK*}
\end{table}

% -------------------------------------------------------
\section{Datasets: WenetSpeech-Formal and Speechio-Formal}
\label{sec:dataset}

We construct WenetSpeech-Formal and Speechio-Formal from the WenetSpeech
corpus~\cite{zhang2022wenetspeech} and Speechio benchmark
data~\cite{speechio2021}, following a two-stage pipeline:
(1) LLM-based formalization:
We firstly collect the verbatim transcripts from WenetSpeech and Speechio, which contain spoken-style characteristics such as filler words, false starts, repetitions, and informal sentence structures. We then prompt DeepSeek-V3.2~\cite{deepseek2024v3} to rewrite each verbatim transcript into formal written Chinese. The rewriting process
includes removing filler words and disfluencies, restructuring
sentences to follow written conventions, normalizing punctuation and
spacing, and correcting obvious recognition errors while preserving
semantic content. The prompt instructs the model to produce concise,
grammatically correct written text that conveys the same meaning as the
spoken input.
(2) Quality filtering:
To improve data quality, we apply an automatic filtering pipeline to remove unreliable rewrites. A sample is excluded if the rewritten text is semantically inconsistent with the source utterance according to embedding-based similarity scores\cite{lahitani2016cosine}, or if the edit distance between the verbatim and formal texts falls outside a predefined range \cite{marzal1993computation}. This process retains only high-quality spoken-to-formal pairs that maintain semantic fidelity while demonstrating a meaningful degree of formalization.

WenetSpeech-Formal contains 969K training samples
derived from the WenetSpeech corpus, covering diverse domains including
audiobooks, podcasts, news broadcasts, and conversational speech.
Speechio-Formal consists of 41K test samples spanning 27 domain-specific
subsets, including lecture recordings, interviews,
meeting transcripts, and spontaneous dialogue, providing a comprehensive
cross-domain evaluation benchmark.
Each sample consists of an audio file paired with two text fields:
\texttt{original\_text} (verbatim transcription) and \texttt{target\_text} (formal written
text produced by DeepSeek-V3.2). Table~\ref{tab:data_examples} shows
representative examples illustrating the types of formalization performed,
including filler-word removal, error correction, and sentence restructuring.

% -------------------------------------------------------
\section{Method}
\label{sec:method}

Given an input audio utterance $\mathbf{x}$, our objective is to directly
predict a formal written transcription $\hat{y}$ in a single pass:
\begin{equation}
  \hat{y} = \arg\max_{y} P_{\theta}(y \mid \mathbf{x}),
\end{equation}
where $y$ denotes a well-formed written sentence rather than a verbatim
spoken transcript. Unlike the conventional two-stage ASR--LLM pipeline,
this formulation couples acoustic recognition and linguistic
formalization into one conditional generation process, so no auxiliary
rewriter is required at inference time.

We instantiate $P_{\theta}(y \mid \mathbf{x})$ with Qwen3-ASR~\cite{qwen3asr2025}
at the 0.6B and 1.7B scales, which adopts a Whisper-style audio encoder\cite{radford2023whisper}
and an autoregressive Qwen decoder\cite{xu2025qwen3omni}. For each training sample, the model
receives speech features extracted from the waveform and is supervised to
generate the corresponding formal target text from
WenetSpeech-Formal and Speechio-Formal. The training objective is standard
teacher-forced maximum likelihood:
\begin{equation}
  \mathcal{L}_{\text{SFT}} = -\sum_{t=1}^{T} \log P_{\theta}(y_t \mid \mathbf{x}, y_{<t}),
\end{equation}
where $T$ is the target length and $y_t$ is the $t$-th token in the formal
reference. Optimizing this objective encourages the model to jointly learn:
acoustic-to-text alignment, disfluency removal, spoken-to-formal
style transfer, and content-preserving rewriting under a unified
end-to-end framework\footnote{Our code is available at \url{https://github.com/TaurenMountain/FormalASR}}.
% Our training details are given in Appendix~\ref{sec:training-details}.

During inference, decoding is performed directly on audio input to produce
formal text, which keeps system complexity identical to a standard single-model
ASR deployment while delivering spoken-to-formal outputs.

% -------------------------------------------------------
\section{Experiments}
\label{sec:experiments}
We evaluate on WenetSpeech-Formal with 31,932 in-domain test samples and Speechio-Formal with cross-domain 41,893 samples.
We report CER ($\downarrow$), ROUGE-L ($\uparrow$), and BERTScore F1 ($\uparrow$) to jointly assess surface-level accuracy and semantic fidelity.
Baselines are Qwen3-ASR-0.6B and Qwen3-ASR-1.7B~\cite{qwen3asr2025} without fine-tuning, and Whisper large-v3~\cite{radford2023whisper} as a cross-system reference.

\begin{table}[t]
\centering
\caption{Spoken-to-formal ASR results on WenetSpeech-Formal and Speechio-Formal benchmarks.}
\label{tab:wenetspeech}
\begin{tabular}{lccc}
\toprule
WenetSpeech-Formal & CER $\downarrow$ & ROUGE-L $\uparrow$ & BERTScore $\uparrow$ \\
\midrule
Qwen3-ASR-0.6B & 0.2581 & 0.8463 & 0.9198 \\
FormalASR-0.6B (Ours) & \textbf{0.1770} & \textbf{0.8769} & \textbf{0.9359} \\
Qwen3-ASR-1.7B & 0.2460 & 0.8571 & 0.9268 \\
FormalASR-1.7B (Ours) & \textbf{0.1606} & \textbf{0.8896} & \textbf{0.9439} \\
Whisper large-v3 & 0.3631 & 0.7393 & 0.8538 \\
\midrule
\midrule
Speechio-Formal & CER $\downarrow$ & ROUGE-L $\uparrow$ & BERTScore $\uparrow$ \\
\midrule
Qwen3-ASR-0.6B & 0.2252 & 0.8701 & 0.9343 \\
FormalASR-0.6B (Ours) & \textbf{0.1603} & \textbf{0.8948} & \textbf{0.9481} \\
Qwen3-ASR-1.7B & 0.2393 & 0.8510 & 0.9108 \\
FormalASR-1.7B (Ours) & \textbf{0.1499} & \textbf{0.9029} & \textbf{0.9533} \\
Whisper large-v3 & 0.3302 & 0.7643 & 0.8795 \\
\bottomrule
\end{tabular}
\end{table}

\begin{figure}[t]
  \centering
  \includegraphics[width=\linewidth]{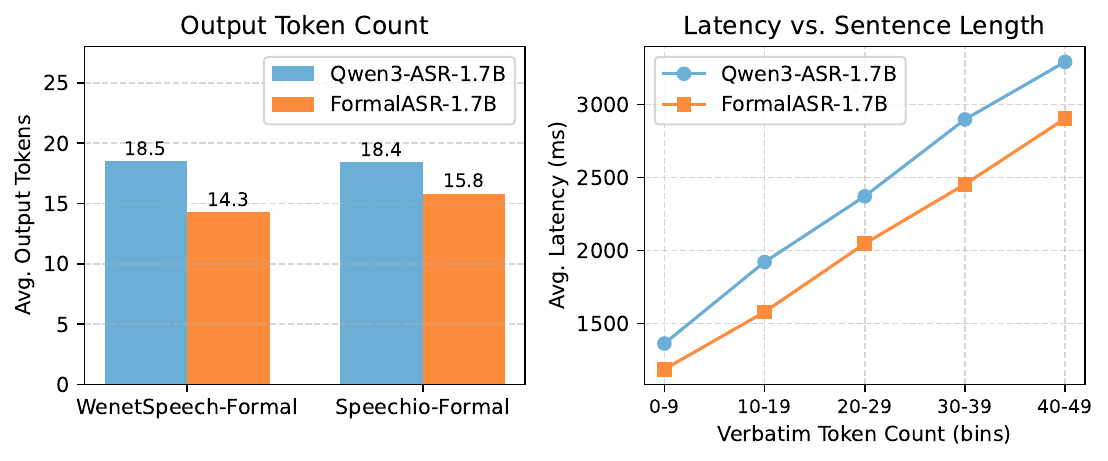}
  \caption{Inference efficiency of FormalASR-1.7B compared with Qwen3-ASR-1.7B (verbatim).
           \textbf{Left}: average output token counts on WenetSpeech-Formal (18.5 for Qwen3-ASR-1.7B, 14.3 for FormalASR-1.7B)
            and Speechio-Formal (18.5 for Qwen3-ASR-1.7B, 15.8 for FormalASR-1.7B), showing that spoken-to-formal conversion
           consistently produces shorter sequences.
           \textbf{Right}: per-sample latency as a function of verbatim sentence
           length (token count bins); FormalASR-1.7B achieves lower latency across
           all sentence lengths, with the gap widening for longer utterances.}
  \label{fig:speed}
\end{figure}

\textbf{Main Results:}
As seen in Table~\ref{tab:wenetspeech}, FormalASR-0.6B and FormalASR-1.7B consistently outperform their verbatim Qwen3-ASR counterparts across all three metrics on both benchmarks, while Whisper large-v3 lags behind all Qwen3-ASR variants due to its verbatim transcription design.
On the in-domain WenetSpeech-Formal benchmark, FormalASR-0.6B reduces CER from 0.2581 to 0.1770, a 31.4\% relative reduction, and FormalASR-1.7B reduces CER from 0.2460 to 0.1606, a 34.7\% relative reduction. ROUGE-L and BERTScore improve in tandem, confirming that the gains are not merely an artifact of shorter outputs.
On the cross-domain Speechio-Formal benchmark, both models maintain strong performance across 27 domain-specific subsets unseen during training: FormalASR-0.6B achieves a 28.8\% relative CER reduction from 0.2252 to 0.1603, and FormalASR-1.7B achieves 37.4\% from 0.2393 to 0.1499, demonstrating that the spoken-to-formal capability learned from WenetSpeech-Formal transfers broadly to diverse speech domains.

\begin{table}[t]
\centering
\caption{GGUF Quantization results of FormalASR on WenetSpeech-Formal.}
\label{tab:quant-gguf}
\begin{tabular}{lccccc}
\toprule
Model & Precision & Model Size & CER  & ROUGE-L & BERTScore \\
\midrule
\multirow{3}{*}{0.6B} & BF16   & 1.46 GB  & 0.1770 & 0.8769 & 0.9359 \\
 & Q8\_0 & 0.78 GB  & 0.1775 & 0.8766 & 0.9357 \\
 & Q4\_K & 0.42 GB  & 0.1969 & 0.8627 & 0.9281 \\
\midrule
\multirow{3}{*}{1.7B} & BF16   & 3.80 GB  & 0.1606 & 0.8896 & 0.9439 \\
 & Q8\_0 & 2.03 GB  & 0.1607 & 0.8896 & 0.9438 \\
 & Q4\_K & 1.08 GB & 0.1744 & 0.8805 & 0.9392 \\
\bottomrule
\end{tabular}
\end{table}

\textbf{Inference Efficiency:}
A side benefit of spoken-to-formal conversion is that removing filler words and disfluencies shortens the output sequence, reducing autoregressive decoding steps. Figure~\ref{fig:speed} compares FormalASR-1.7B with the verbatim Qwen3-ASR-1.7B baseline.
The left panel shows average output token counts. On WenetSpeech-Formal, FormalASR-1.7B generates 14.3 tokens per utterance versus 18.5 for Qwen3-ASR-1.7B, a 22.8\% reduction. On Speechio-Formal, output length decreases from 18.5 to 15.8 tokens, corresponding to a 14.3\% reduction.
The right panel shows decoding latency across five sentence-length bins. For short utterances with 0--9 tokens, FormalASR-1.7B decodes in about 1,188ms compared with 1,364ms for Qwen3-ASR-1.7B. The latency gap increases with length: in the 20--29 token bin, FormalASR-1.7B is about 324ms faster, while in the 40--49 token bin the gap reaches about 388ms.
The growing latency advantage stems from the fact that longer verbatim transcripts contain more disfluencies, allowing spoken-to-formal conversion to remove more tokens and reduce decoding steps. This makes FormalASR especially beneficial for long-form speech such as meetings and lectures.

\textbf{Quantization:}
To further validate the on-device motivation, we evaluate post-training
quantization on FormalASR checkpoints using the GGUF format~\cite{gerganov2023llamacpp},
which is widely supported by on-device inference.
We quantize both FormalASR-0.6B and FormalASR-1.7B to 8-bit (Q8\_0) and
4-bit (Q4\_K) precision and measure the resulting accuracy and model-size
trade-offs on the WenetSpeech-Formal testset.
As shown in Table~\ref{tab:quant-gguf}, Q8\_0 quantization is nearly
lossless, reducing model size by about 47\% while leaving CER,
ROUGE-L, and BERTScore essentially unchanged for both models.
More aggressive Q4\_K quantization achieves around 72\% compression,
shrinking FormalASR-1.7B from 3.80\,GB to 1.08\,GB and FormalASR-0.6B
from 1.46\,GB to 0.42\,GB, with moderate quality degradation.
Despite the accuracy drop, the 1.08\,GB FormalASR-1.7B Q4\_K model
still outperforms the 1.46\,GB FormalASR-0.6B BF16 model across all
metrics, demonstrating that FormalASR maintains strong spoken-to-formal
performance under practical on-device deployment constraints.

% -------------------------------------------------------
\section{Conclusion}
\label{sec:conclusion}
We presented two contributions toward end-to-end spoken-to-formal Chinese ASR.
First, we constructed and open-sourced WenetSpeech-Formal and Speechio-Formal,
the first large-scale datasets for spoken-to-formal Chinese transcription.
Second, we fine-tuned FormalASR, compact end-to-end models at the 0.6B and
1.7B scales that directly transcribe spoken Chinese into formal written text
without any auxiliary LLM at inference time.
FormalASR achieves up to 37.4\% relative CER reduction over verbatim
baselines, with consistent ROUGE-L and BERTScore improvements across
in-domain and cross-domain benchmarks, while also reducing decoding
latency through shorter output sequences. Furthermore, GGUF quantization
demonstrates strong deployability, preserving most accuracy under
substantially reduced memory footprints.

\bibliographystyle{IEEEtran}
\bibliography{ref}

\end{document}